\documentclass{article}

\usepackage{microtype}
\usepackage{graphicx}
\usepackage{subcaption}
\usepackage{booktabs} 

\usepackage{amsfonts}
\usepackage{multirow}
\usepackage{colortbl}
\usepackage{float}
\usepackage[T1]{fontenc}
\usepackage{lmodern}
\usepackage{xspace}
\usepackage[table]{xcolor}

\usepackage{hyperref}

\usepackage[preprint]{icml2026}

\usepackage{amsmath}
\usepackage{amssymb}
\usepackage{mathtools}
\usepackage{amsthm}

\usepackage[capitalize,noabbrev]{cleveref}

\theoremstyle{plain}

\theoremstyle{definition}

\theoremstyle{remark}

\newcommand{\method}{{\fontfamily{lmtt}\selectfont\textsc{OmniVideo-R1}}\xspace}

\usepackage[textsize=tiny]{todonotes}

\icmltitlerunning{\method}

\begin{document}
\twocolumn[
  \icmltitle{
  \method: Reinforcing Audio-visual Reasoning with \\ Query Intention and Modality Attention}




  \icmlsetsymbol{mark}{*}
  \begin{icmlauthorlist}
    \icmlauthor{Zhangquan Chen$^*$}{thu}
    \icmlauthor{Jiale Tao$^\dagger$}{tencent}
    \icmlauthor{Ruihuang Li}{tencent}
    \icmlauthor{Yihao Hu}{hnu}
    \icmlauthor{Ruitao Chen}{tencent}
    \icmlauthor{Zhantao Yang}{tencent}
    \icmlauthor{Xinlei Yu}{nus}
    \icmlauthor{Haodong Jing}{xjtu}
    \icmlauthor{Manyuan Zhang}{cuhk}
    \icmlauthor{Shuai Shao}{tencent}
    \icmlauthor{Biao Wang}{tencent}
    \icmlauthor{Qinglin Lu}{tencent}
    \icmlauthor{Ruqi Huang$^\dagger$}{thu}
  \end{icmlauthorlist}

  \icmlaffiliation{thu}{THU}
  \icmlaffiliation{tencent}{Tencent HY}
  \icmlaffiliation{nus}{NUS}
  \icmlaffiliation{cuhk}{CUHK}
  \icmlaffiliation{hnu}{HNU}
  \icmlaffiliation{xjtu}{XJTU}

  \icmlcorrespondingauthor{Ruqi Huang}{ruqihuang@sz.tsinghua.edu.cn}
  \icmlcorrespondingauthor{Jiale Tao}{jialetao.std@gmail.com}

  \icmlkeywords{Machine Learning, ICML}

  \vskip 0.3in

]



\printAffiliationsAndNotice{$^*$ The work was conducted during the internship of Zhangquan Chen (czq23@mails.tsinghua.edu.cn) at Tencent HY.}  
\begin{abstract}
While humans perceive the world through diverse modalities that operate synergistically to support a holistic understanding of their surroundings,
existing omnivideo models still face substantial challenges on audio-visual understanding tasks.
In this paper, we propose \method, a novel reinforced framework that improves mixed-modality reasoning. 
\method empowers models to ``think with omnimodal cues'' by two key strategies: (1) query‑intensive grounding based on self‑supervised learning paradigms; and (2) modality‑attentive fusion built upon contrastive learning paradigms.
Extensive experiments on multiple benchmarks demonstrate that \method consistently outperforms strong baselines, highlighting its effectiveness and robust generalization capabilities.

\end{abstract}
\begin{figure}[!htbp]
    \centering
    \includegraphics[width=0.5\textwidth]{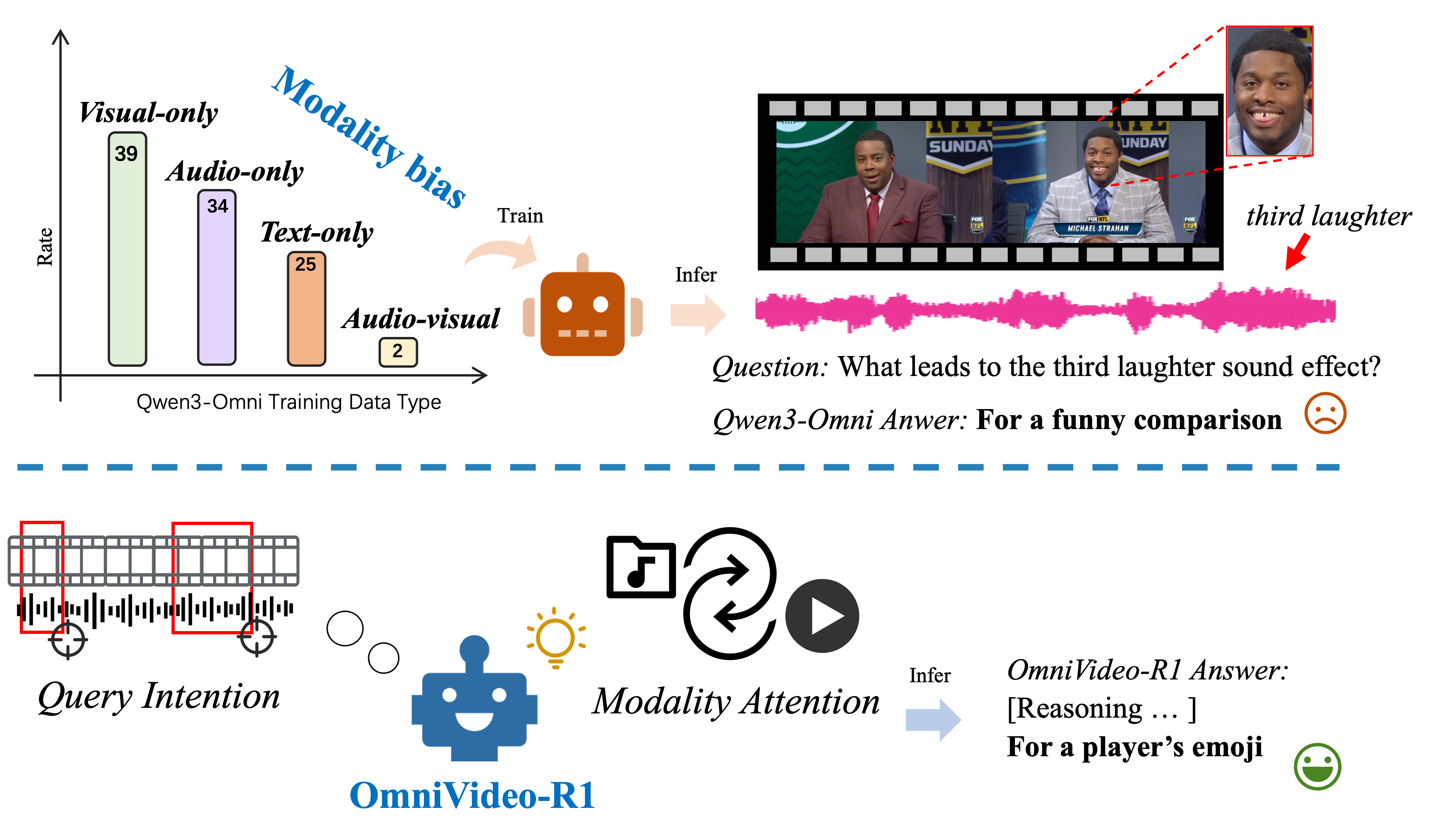}
    \caption{Pre-trained MLLMs (e.g., Qwen3-Omni) often exhibit suboptimal performance in audio-visual reasoning tasks due to inherent modality bias. To address this limitation, we reinforce the audio-visual reasoning ability by leveraging query intention and modality attention.}\label{fig:teaser}
    \vspace{-1.2 em}
\end{figure}
\section{Introduction}
\label{sec:intro}
Human cognition is inherently multimodal; we perceive the physical world by processing visual and auditory signals in parallel, integrating them to construct a coherent understanding of complex environments~\cite{zhou2025dailyomni, worldsense, zhao2025tartan,  chen2024three, lu2025predicting}. As Large Language Models (LLMs) evolve into Multimodal LLMs (MLLMs), the ability to interpret such multisensory inputs has become a cornerstone of artificial general intelligence~\cite{yu2025vismem, yu2025visualmultiagent, cai2025does, 10800533, li2025catch}. However, contrary to the expectation that more modalities yield better understanding, current omnimodal models often exhibit a paradoxical behavior.

This phenomenon is also evident in the state-of-the-art models. 
As shown in Fig.~\ref{fig:teaser}, pre-training inherently involves trade-offs across heterogeneous tasks, which can induce a natural modality bias. Consequently, within the Qwen3-30B-A3B family, the Omni variant~\cite{Qwen3-Omni} (audio-visual) substantially underperforms the VL variant~\cite{Qwen3-VL} (visual-only), dropping from 72.1 to 68.5 on MMStar~\cite{mmstar} and from 80.1 to 75.9 on MathVista\_{mini}~\cite{lu2023mathvista}.
These results expose a key limitation of current paradigms: instead of enabling synergistic fusion, \emph{incorporating the audio modality can undermine the model’s established visual reasoning capability}.

A natural response is to increase mixed audio-visual supervision during pre-training; however, scaling high-quality mixed-modality data and aligning it with downstream reasoning needs is non-trivial. On the other hand, existing post-training pipelines commonly rely on supervised fine-tuning (SFT) or vanilla reinforcement learning (RL) (e.g., GRPO)~\cite{zhao2025humanomni, xing2025echoink, yang2025humanomniv2, zhang2023videollama, sun2024videosalmonn}.
Yet these post-training methods do not explicitly train \emph{audio-visual mixed-modality reasoning} behaviors, such as locating and composing evidence across modalities. That is, they provide little supervision over intermediate evidence-tracking. As a result, the model may \emph{ignore decisive audio or visual cues and still produce the correct answer by exploiting dataset biases or unimodal shortcuts}.

To address this challenge, we present \method, 
the first post-training framework designed to improve mixed-modality reasoning.
We posit that solving such problem requires more than just balancing datasets; it requires instilling robust reasoning behaviors that allow the model to actively select and fuse information. Specifically, \method optimizes two fundamental capabilities: (1) \textbf{query-intensive grounding} and (2) \textbf{modality-attentive fusion} built upon query-intensive reasoning.

Inspired by the “think with images” paradigm~\cite{openai2025o3o4mini}, we first introduce \emph{query-intensive grounding}, which enables the model to explicitly localize and reason over audio-visual segments relevant to the user’s query before generating a response. 
Since the grounding annotations conditioned on query intent are costly to obtain, we design a \emph{self-supervised training scheme} that leverages multiple time–caption pairs. This design allows the model to generate grounding hypotheses and validate them against the corresponding textual descriptions.

For learning a robust query intention behavior, we then propose \emph{modality-attentive fusion}, which maximize the utilization of audio-visual cues. To achieve this, we design a \emph{contrastive learning-based strategy} that explicitly encourages the model to derive higher confidence from mixed audio-visual inputs compared to single-modality counterparts. This forces the model to discover synergistic relationships between visual and audio events, ensuring that the fused representation is strictly superior to its constituent parts.

By combining these strategies into a unified RL framework, 
\method turns mixed-modality understanding into a query-driven reasoning process with audio-visual cues.

Our primary contributions are summarized as follows:
\begin{itemize}
    \item We propose \method, the first RL-based framework designed to improve mixed-modality reasoning.
    
    \item We construct a high-quality corpus of 80K audio–visual training samples through a dedicated data-cleaning pipeline, specifically curated to support complex reasoning tasks.
    
    \item We introduce a two-stage RL paradigm that incorporates \emph{self-supervised grounding} and \emph{contrastive fusion}, enabling the model to learn query intention and modality attention without relying on process-level annotations.
    
    \item Extensive experiments demonstrate that \method consistently outperforms strong open-source baselines on audio-visual benchmarks while effectively maintaining robust visual-only performance.
\end{itemize}
\section{Related Work}
\label{sec:related}
\begin{figure*}[!htbp]
    \centering
    \vspace{-0.3em}
    \includegraphics[width=\textwidth]{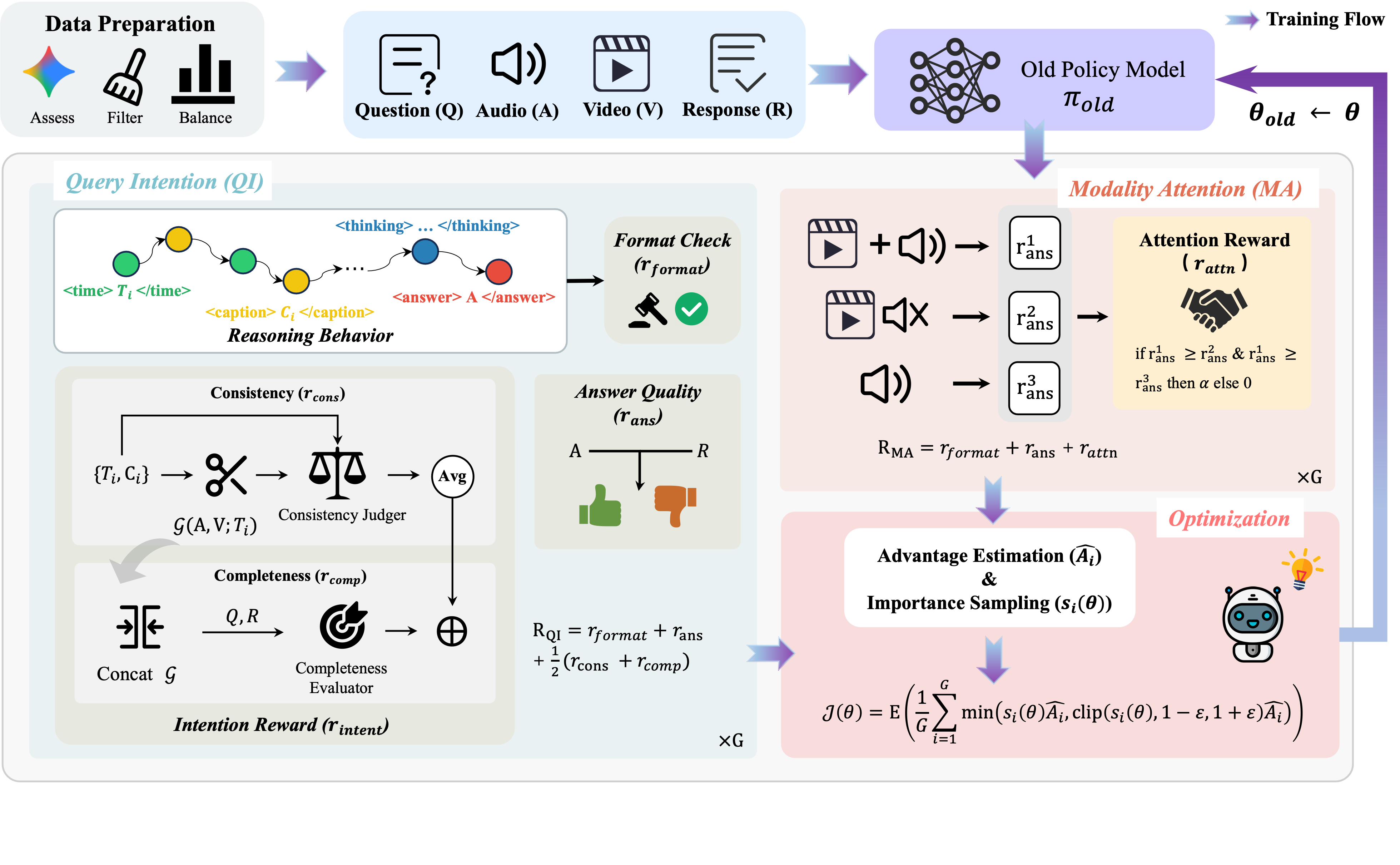}
    \vspace{-3.5em}
    \caption{The schematic illustration of our \method. Based on the dataset collected from data preparation, our training consists of two stages: (1) QI stage establishes query-intensive grounding behavior by aligning multiple time–caption pairs without process-level annotations. (2) MA stage further performs modality-attentive fusion by optimizing a contrastive modality reward.}\label{fig:pipeline}
\end{figure*}
\subsection{Omnimodal Large Language Models}
The integration of audio and visual modalities is more closely to real-world recordings~\cite{liu2023siamhas, deng2024separation, wang2025siamctca, zhao2023benchmark, wang2025fuzzy, diffusionsar}, and requires models to form a cohesive understanding of the surroundings, like humans~\cite{zhao2025humanomni}. Early efforts always focused on silent video understanding~\cite{bai2025qwen2.5, zhang2024video, feng2025video} or treated audio a simple add-on to text~\cite{li2025videochat}, which fragments natural omnimodal representations and thereby limits performance. 

Consequently, subsequent works have aimed for deeper cross-modal fusion. MiniCPM-o-2.6~\cite{yao2024minicpm} and Baichuan-Omni-1.5~\cite{li2024baichuan} extend vision–language foundations with audio processing capabilities, enabling operation across a broader range of modalities. Ola~\cite{liu2025ola} adopts a progressive modality-alignment strategy that incrementally strengthens the language model’s ability to exploit additional modalities. The Video-LLaMA series~\cite{zhang2023videollama} concatenates audio and visual tokens to support joint audio–video understanding, whereas the Video-SALMONN~\cite{sun2024videosalmonn} series employs a multi-resolution causal Q-Former~\cite{li2023blip} to process audio and video simultaneously. Moreover, InternVideo series~\cite{wang2022internvideo} aligns video with audio events, speech, and text via cross-modal contrastive learning, thereby facilitating integrated audio–video representation learning. Qwen2.5-Omni~\cite{xu2025qwen2} introduces a “thinker–talker” architecture, an end-to-end multimodal framework capable of perceiving diverse input types. More recently, Qwen3-Omni~\cite{Qwen3-Omni} leverages an Audio Transformer (AuT) for audio encoding and incorporates TM-RoPE, further enhancing the audio–visual understanding capabilities. 

Despite these advances, current multimodal models \emph{still exhibit substantial limitations on complex tasks}, particularly in scenarios that demand tightly \emph{integrated audio–visual understanding} or more \emph{sophisticated logical reasoning}.
\subsection{Reinforced Multimodal Reasoning}
Reinforcement learning has become a widely adopted paradigm for enhancing the performance of large language models~\cite{offlinerl, lan2025mappo}. Recent work combines RL with vision and language to elicit stronger reasoning capabilities~\cite{chen2025think, ni2025recondreamer}. Some approaches, inspired by DeepSeek-R1~\cite{rela:deepseekr1}, introduce purely textual chain-of-thought~\cite{rela_new:llamavo1, rela_new:r1v, rela:insight-v}. Building on this, methods such as VisRL~\cite{chen2025visrl}, SIFThinker~\cite{chen2025sifthinker}, GRIT~\cite{fan2025grit}, and CogCoM~\cite{qi2024cogcom} enable “thinking with images” by integrating visual evidence into the reasoning trajectory.

Beyond these efforts, several studies have extended the notion of reasoning to omnimodal models. R1-Omni~\cite{zhao2025r1} is primarily designed for audio–visual referring segmentation, whereas EchoInk-R1~\cite{xing2025echoink} investigates the direct application of vanilla GRPO~\cite{rela:deepseekr1} in the omnimodal setting. In addition, Omni-R1~\cite{zhong2025omni} adopts a dual-system architecture to tackle long-horizon video–audio reasoning, and HumanOmnv2~\cite{yang2025humanomniv2} enhances model capabilities through training on datasets curated for complex human intention understanding.

However, compared with silent video reasoning~\cite{video-r1, wang2025videothinker}, \emph{current explorations of omnimodal reasoning remain relatively limited}. 
Existing approaches concentrate on directly transferring vanilla RL, designing intricate multi-branch architectures, or constructing specialized training datasets. Yet omnimodal models \emph{inherently require deeper multimodal fusion in order to unlock stronger reasoning capabilities}, thereby achieve genuine “aha moments.” 
Consequently, there is still a conspicuous absence of training methodologies that are tailored to the distinctive characteristics of such omnimodal models.

\section{Methodology}
\label{sec:method}
\paragraph{Preliminary.} Reinforcement learning~\cite{christiano2017deep} has emerged as a particularly effective approach for substantially enhancing the robustness and factual accuracy of large language models~\cite{ouyang2022training}. In practice, off-policy learning settings are typically used during policy model training to improve sample efficiency. However, for Mixture-of-Experts (MoE) models (e.g., Qwen3-Omni-30B-A3B~\cite{Qwen3-Omni}), the activation of different experts can induce substantial shifts in the token distribution. Under such conditions, token-level importance sampling often introduces high-variance noise into the training gradients, which accumulates over long sequences and is further exacerbated by clipping mechanisms. To this end, our method performs optimization directly at the sequence level, following the formulation introduced by Group Sequence Policy Optimization (GSPO) algorithm~\cite{zheng2025group}. The optimization objective
is formulated as:

\begin{equation}
\mathcal{J}(\theta) = \mathbb{E}_{x \sim \mathcal{D}, \{y_i\}_{i=1}^G \sim \pi_{\theta_{\text{old}}}(\cdot|x)} (P_\theta),
\end{equation}

where the response $y_i$ is sampled from old policy model $\pi_{\theta_{\text{old}}}$ based on the input $x$, and $P_\theta$ is:
\begin{equation}
P_\theta = \frac{1}{G} \sum_{i=1}^G \min \left( s_i(\theta) \widehat{A}_i, \text{clip} \left( s_i(\theta), 1 - \varepsilon, 1 + \varepsilon \right) \widehat{A}_i \right).
\end{equation}

Here, we also adopt the group-based advantage estimation:
\begin{equation}
    \widehat{A}_i = \frac{R(x, y_i) - \text{mean} \left( \{R(x, y_i)\}_{i=1}^G \right)}{\text{std} \left( \{R(x, y_i)\}_{i=1}^G \right)},
\end{equation}

$R(\cdot)$ denotes the reward function that will be introduced below, and we define the importance ratio $s_i(\theta)$ based on sequence likelihood~\cite{zheng2023click}:
\begin{equation}
    s_i(\theta) = \exp \left( \frac{1}{|y_i|} \sum_{t=1}^{|y_i|} \log \frac{\pi_\theta(y_{i,t} | x, y_{i,<t})}{\pi_{\theta_{\text{old}}}(y_{i,t} | x, y_{i,<t})} \right).
\end{equation}

\begin{figure}[!htbp]
    \centering
    \includegraphics[width=0.45\textwidth]{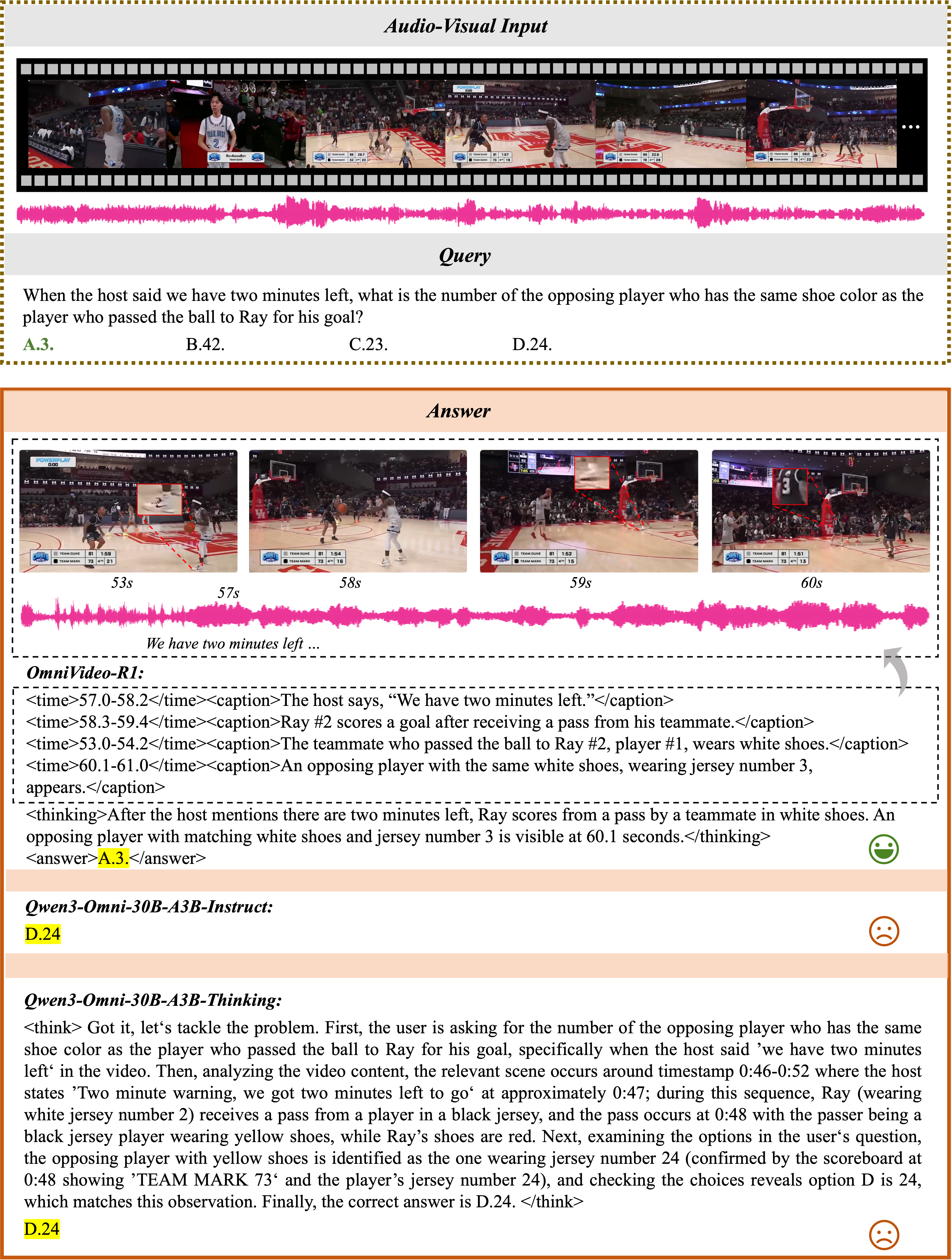}
    \caption{Visualization of the responses and underlying reasoning process generated by \method and Qwen3-Omni-30B-A3B-Instruct, -Thinking to an audio-visual understanding question.}\label{fig:visual}
    \vspace{-1.0em}
\end{figure}
\paragraph{Method overview.} As shown in Fig.~\ref{fig:pipeline}, \method adopts GSPO to optimize the entire reasoning process, enabling the model to extract intention-relevant cues and to effectively integrate audio–visual information throughout reasoning. This model behavior emerges through two training stages. That is, we first induce the model to develop query-intensive reasoning behavior, and then, further train it to integrate multiple modalities in a logically consistent manner. In the first stage (QI), the model is trained with a \emph{self-supervised objective} defined over multiple pairs of grounding and caption generated within the reasoning trajectory. 
In the second stage (MA), we promote deeply fused understanding by first decoupling the modality-specific inputs and then performing \emph{contrastive learning} across them. Notably, throughout the entire training pipeline, \method doesn't rely on any explicit process-level annotations for query-intensive grounding or modality fusion. 

Fig.~\ref{fig:visual} illustrates our reasoning process in comparison with Qwen3-Omni-30B-A3B. Our \method endows the model with the ability to "think with omnimodal cues", i.e., to perform query-intensive grounding that identifies key cues, thereby enabling more accurate and reliable reasoning to the final answer.

\subsection{Data Preparation}
\label{sec:data}
We first collect raw data from LLaVA-Video~\cite{zhang2024video} and Video-Vista~\cite{li2024videovista}, and perform structural validation to remove problematic samples with metadata issues (e.g., silent videos). To further filter out l viow-quality samples that are misaligned with our multimodal setting, we apply a three-stage refinement pipeline as shown in Fig.~\ref{fig:supdata}, which consists of (i) quality assessment, (ii) heuristic filtering, and (iii) categorical balancing.
\begin{figure}[!htbp]
    \centering
    \includegraphics[width=0.48\textwidth]{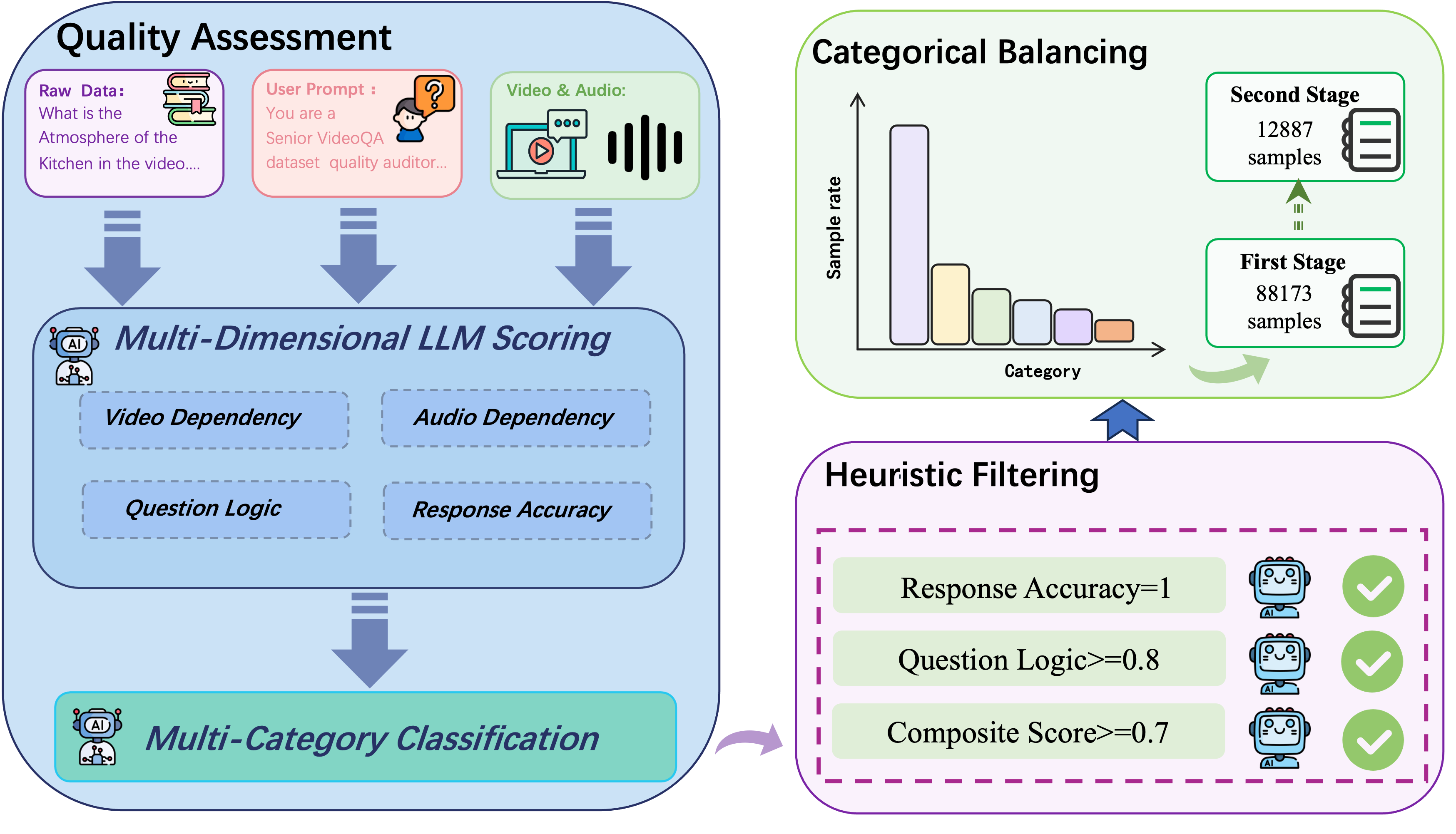}
    \caption{Pipeline for our data preparation consisting of 3 stages.}\label{fig:supdatappl}
\end{figure}

For data quality assessment, we employ Gemini-2.5-Pro~\cite{google_gemini_25_pro} to score each sample along four key dimensions: video dependency $s_v$, audio dependency $s_a$, question logic $s_q$, and response accuracy $s_r$. Each dimension is normalized to a maximum of 1, and the weighted composite score $s_c$ is computed as a weighted sum over the four dimensions. Subsequently, Qwen-3-32B~\cite{yang2025qwen3} is used to categorize the samples according to the 15-category taxonomy described in the Appendix.

After scoring and categorizing data, we apply the following filtering rules: (i) $s_r = 1$, (ii) $s_q \ge 0.8$, (iii) $s_c \ge 0.7$. Any sample that fails to satisfy any of these rules is discarded.

Finally, we mitigate long-tail bias by pruning sparse categories (i.e., those with fewer than 10 samples). Observing a significant gap between the top two classes, we further require that the number of samples in the largest class does not exceed three times that of the second-largest class. Specifically, we first retain all samples with both $s_a = 1$ and $s_i = 1$, then sort the remaining data in descending order of $s_c$ and remove samples that exceed the specified count, resulting in a smoother data distribution.

After applying all of the above steps, we obtain 88173 examples for the first training stage training. Considering the high-quality requirements for audio-visual fused data in the second training stage, we then derive a subset of 12887 examples by keeping only instances with high audio-visual dependency, i.e., $s_v \ge 0.7$ and $s_a \ge 0.7$.

\subsection{Query-intensive Grounding (QI)}
\label{sec:stage1}
\begin{table*}[t!]
    \centering
    \footnotesize
    \setlength{\tabcolsep}{3.5mm}
    \begin{tabular}{lcccc}
    \hline
    \textbf{Method}             & \textbf{Daily-Omni} & \textbf{WorldSense} & \textbf{IntentBench} & \textbf{VideoHolmes} \\ \hline
    \multicolumn{5}{c}{\textit{Closed-source Models}}                                                                     \\
    Gemini-2.0-Flash            & 67.8                & 56.2                & 67.8                 & 30.6                 \\
    Gemini-2.5-Pro~\cite{google_gemini_25_pro}              & 81.4                & 64.6                & 67.2                 & 64.9                 \\
    Gemini-3-Pro                & 81.1                & 66.4                & 71.5                 & 67.0                 \\ \hline
    \multicolumn{5}{c}{\textit{Open-source Models}}                                                                       \\
    VideoLLaMA2-7B~\cite{cheng2024videollama2}              & 35.2                & 25.4                & —                    & 35.2                 \\
    Qwen2.5-Omni-7B~\cite{qwenomni2.5}             & 47.5                & 45.4                & 64.2                 & 16.4                 \\
    MiniCPM-o-7B~\cite{yao2024minicpm}                & 53.1                & —                   & 54.5                 & —                    \\
    VITA-1.5-7B~\cite{vita-1.5}                 & —                   & 36.9                & 54.2                 & —                    \\
    Ola-7B~\cite{liu2025ola}                      & 49.9                & —                   & 57.4                 & —                    \\
    HumanOmniV2-7B~\cite{yang2025humanomniv2}              & 58.5                & 47.1                & 69.3                 & —                    \\
    video-SALMONN 2+-7B~\cite{video-salmonn-2}         & 71.8                & 50.9                & —                    & 46.9                 \\
    video-SALMONN 2+-72B~\cite{video-salmonn-2}        & 79.4                & 56.5                & —                    & 57.8                 \\
    Qwen3-Omni-30B-A3B-Instruct~\cite{Qwen3-Omni}          & 63.6                & 54.0                & 65.7                 & 49.0                 \\
    Qwen3-Omni-30B-A3B-Thinking~\cite{Qwen3-Omni} & 75.8                & 48.0                & 68.5                 & 57.3                 \\
    \rowcolor[HTML]{E7E6E6} 
    \rowcolor[HTML]{E7E6E6} 
    \method & \textbf{82.8}       & \textbf{65.8}       & \textbf{74.2}        & \textbf{62.9}        \\ \hline
    \end{tabular}
    \caption{Performance of different methods on a range of audio-visual benchmarks, including Daily-Omni~\cite{dailyomni}, WorldSense~\cite{worldsense}, IntentBench~\cite{yang2025humanomniv2}, and VideoHolmes~\cite{videoholmes}. Our training was conducted on QI and on both QI + MA. The \textbf{best} is highlighted, and the second-best is \underline{underlined}.}\label{table:omni}
\end{table*}
Query-intensive grounding operations aim to help the model identify key frames containing critical audio-visual cues within a video sequence~\cite{wang2025videothinker}. However, human annotations of prompt-related key frames are often complex and expensive. To address this issue, we propose a novel grounding approach that establishes a correspondence between grounding and captioning \emph{without relying on any dense annotations}, thereby enabling \emph{self-supervised learning of the model’s procedural behavior}.

Specifically, given one question and the corresponding audio–visual content $(Q, A, V)$, we encourage the model to produce outputs in the structured format {\small\texttt{<time>...</time><caption>...</caption> ... <thinking>...</thinking><answer>...</answer>}}. A reward of $r_{\mathrm{format}} = 1.0$ is assigned to responses that strictly comply with this output template. For each rollout, we denote the multiple generated time–caption pairs by $\{T_1, C_1, T_2, C_2, \ldots, T_N, C_N\}$.
We then perform self-supervised learning by evaluating the consistency reward between each $T_i$ and $C_i$, i.e.,
\begin{equation}
r_{\mathrm{cons}} = \frac{1}{N} \sum_{i=1}^{N} E_{\mathrm{cons}}^{(L)}\bigl(\mathcal{G}(A, V; T_i),\, C_i\bigr),
\end{equation}
where $\mathcal{G}(A, V; T_i)$ extracts the audio-visual segment from $(A, V)$ corresponding to the time span $T_i$, and $E_{\mathrm{cons}}^{(L)}(\cdot)$ denotes a soft evaluation function implemented via a judger model (i.e., Qwen3-VL-235B-A22B-Instruct) with $L$ predefined rules.
The detailed rules and the associated prompts are provided in Appendix.

On the one hand, we perform self-supervised learning by enforcing the correctness of each time–caption pair. On the other hand, we also require the grounding to be precise, i.e., it should (i) minimally and effectively cover all ground-truth intention-related cues \(T_{\mathrm{gt}}\), and (ii) avoid redundant predictions. Formally, for each $i,j \leq N$, we except:
\begin{equation}
    \label{eq:time}
    \left[
        \left( \bigcup_{i=1}^{N} T_i \right) \cap T_{\mathrm{gt}} = T_{\mathrm{gt}}
    \right]
    \;\land\;
    \left[
        T_i \cap T_j = \varnothing,\ \forall\, i \neq j
    \right].
\end{equation}

However, in this work, we tackle the challenging setting where no ground-truth \(T_{\mathrm{gt}}\) is available, and instead propose a soft approximation to solve Eq.~\ref{eq:time}. Specifically, we first crop all predicted segments and then concatenate them into a single sequence, which is subsequently evaluated along two dimensions: content completeness and precision. In other words, we assess whether the audio-visual information contained within the grounded segments is \emph{adequate and accurate for supporting the reasoning process from the question \(Q\) to the final answer \(R\)}. Accordingly, we define the completeness reward as: 
\begin{equation}
    r_{\mathrm{comp}}
    = E_{\mathrm{comp}}^{(M)}\Bigl(
        \bigoplus_{i=1}^{N} \mathcal{G}(A, V; T_i),
        \, Q,
        \, R
    \Bigr),
\end{equation}
where \(\bigoplus_{i=1}^{N} \mathcal{G}(A, V; T_i)\) denotes the temporally ordered concatenation of all grounded audio-visual segments, yielding a single composite video clip. Here, \(E_{\mathrm{comp}}^{(M)}(\cdot)\) is the intent evaluation function instantiated with \(M = 3\) predefined rules. More details are listed in the Appendix.

Meanwhile, we also leverage the outcome signal, following~\cite{rela:deepseekr1}. Specifically, we softly evaluate the quality of the final answer and assign a continuous score $r_{\mathrm{ans}}$; the detailed evaluation protocol is provided in Appendix. Finally, the reward in our QI training stage is defined as
\begin{equation}
    R_{\mathrm{QI}}
    = r_{\mathrm{format}} + r_{\mathrm{ans}} + \frac{1}{2}\bigl(r_{\mathrm{cons}} + r_{\mathrm{comp}}\bigr).
\end{equation}

We establish a unified training framework from three complementary perspectives: (i) global format regularization $r_{\mathrm{format}}$, (ii) outcome-based constraints $r_{\mathrm{ans}}$, and (iii) process-level self-supervision $r_{\mathrm{intent}} = \frac{1}{2}\bigl(r_{\mathrm{cons}} + r_{\mathrm{comp}}\bigr)$. Under this training design, the model is expected to \emph{infer the underlying intention, extract task-relevant cues, and perform reasoning over these audio–visual content.}

\subsection{Modality-attentive Fusion (MA)}
\label{sec:stage2}
As QI stage is primarily evaluated in a vision-centric manner, relying solely on query-intensive grounding still prevents the model from capturing the subtle but decisive sound cues (as shown in Fig.~\ref{fig:show1}). This inability to leverage audio cues further leads to substantial redundant outputs (as shown in Fig.~\ref{fig:show2}). 
To address this issue, we propose a \emph{modality-attentive fusion} scheme, whose central idea is to encourage the model to \emph{fully exploit and synergistically integrate both audio and visual information to improve accuracy}.

Concretely, for each input $x$, we compare the model's performance under three rollout settings: (i) combined audio--visual input; (ii) silent-video-only input; and (iii) audio-only input. For a desirable multimodal understanding model, the performance with full multimodal input should not be inferior to that with any single-modality input, especially on datasets where both acoustic and visual cues are required to correctly answer the question. Denote the soft scores associated with these three rollouts by $r_{\text{ans}}^{1}$, $r_{\text{ans}}^{2}$, and $r_{\text{ans}}^{3}$, respectively. We then define the \emph{attention} reward as
\begin{equation}
r_{\mathrm{attn}} =
\begin{cases}
\alpha, & \text{if } r_{\text{ans}}^{1} \ge r_{\text{ans}}^{2} \text{ and } r_{\text{ans}}^{1} \ge r_{\text{ans}}^{3} \\
0, & \text{otherwise}
\end{cases}
\end{equation}
where $\alpha$ is a hyperparameter controlling the magnitude of the \emph{attention} reward (set to $\alpha = 0.3$ in our experiments). This contrastive formulation explicitly encourages the model to achieve superior performance when audio and visual information are effectively fused, rather than relying predominantly on a single modality.

Building upon the \emph{contrastive learning} strategy, our MA training stage focuses on enhancing model capabilities in a more targeted subset of data which specifically requires integrated audio–visual understanding. This stage aims to advance the reasoning paradigm from query-intensive grounding to deeper multimodal understanding. Formally, the reward for this stage is defined as:
\begin{equation}
    R_{\mathrm{MA}}
    = r_{\mathrm{format}} + r_{\mathrm{ans}} + r_{\mathrm{attn}}.
\end{equation}

\section{Experiments}
\label{sec:exp}
We evaluate \method with several state-of-the-art (SOTA) methods on an array of categories as follows. More details about benchmarks are listed in the Appendix.

\noindent\textbf{Training.} \method is trained based on Qwen3-Omni-30B-A3B following the pipeline described in Sec.~\ref{sec:method}. As detailed in Sec.~\ref{sec:data}, we use 88173 samples for QI stage (Sec.~\ref{sec:stage1}) and 12887 samples for MA stage (Sec.~\ref{sec:stage2}).

\noindent\textbf{Hyper-parameters.} For \method, we conduct training under a 128$\times$H20 setup with a global batch size of 256. We set the balancing coefficient of all rewards to 1 and use a learning rate of $1\times10^{-6}$. The rollout number is 8, and the maximum sequence length is 32768. Additional details are provided in the Appendix.

\noindent\textbf{Evaluation metric.} For multiple-choice questions, we report Accuracy, which is calculated based on exact matches between the model’s predictions and the ground truth.
\subsection{Omnimodal Understanding}
\begin{table*}[t!]
    \centering
    \footnotesize
    \setlength{\tabcolsep}{0.4mm}
    \begin{tabular}{lccclccccc}
    \hline
                                      & \multicolumn{3}{c}{\textbf{Audio Type}}       &  & \multicolumn{4}{c}{\textbf{Video Duration}}                   &                                 \\ \cline{2-4} \cline{6-9}
    \multirow{-2}{*}{\textbf{Method}} & Music         & Sound         & Speech        &  & (0,1{]} min   & (1,5{]} min   & (5,10{]} min  & (10,30{]} min & \multirow{-2}{*}{\textbf{Avg.}} \\ \hline
    \multicolumn{10}{c}{\textit{Closed-source Models}}                                                                                                                                     \\
    Gemini-2.0-Flash                  & 29.7          & 40.3          & 43.2          &  & 49.4          & 43.2          & 41.1          & 34.9          & 41.5                            \\
    Gemini-2.5-Pro~\cite{google_gemini_25_pro}                    & 38.5          & 57.7          & 61.7          &  & 57.8          & 64.4          & 55.0          & 55.9          & 58.9                            \\
    Gemini-3-Pro                      & 56.2          & 54.1          & 55.7          &  & 61.0          & 56.4          & 52.9          & 52.5          & 55.5                            \\ \hline
    \multicolumn{10}{c}{\textit{Open-source Models}}                                                                                                                                       \\
    VideoLLaMA2-7B~\cite{cheng2024videollama2}                    & 26.4          & 30.7          & 29.3          &  & 32.0          & 28.2          & 29.6          & 28.3          & 29.2                            \\
    Qwen2.5-Omni-7B~\cite{xu2025qwen2}                  & 23.1          & 25.3          & 30.7          &  & 41.6          & 27.4          & 25.3          & 26.7          & 29.3                            \\
    MiniCPM-o-7B~\cite{yao2024minicpm}                      & 27.5          & 28.6          & 30.2          &  & 31.4          & 28.5          & 34.5          & 26.2          & 29.7                            \\
    HumanOmniV2-7B~\cite{yang2025humanomniv2}                    & 20.9          & 31.1          & 31.6          &  & 36.6          & 29.4          & 29.6          & 29.3          & 30.5                            \\
    Baichuan-Omni-1.5-7B~\cite{li2024baichuan}              & 24.2          & 31.3          & 31.4          &  & 28.9          & 31.8          & 28.4          & 32.4          & 30.7                            \\
    Qwen3-Omni-30B-A3B-Instruct~\cite{Qwen3-Omni}                & 30.8          & 35.8          & 38.0          &  & 46.8          & 37.4          & 37.7          & 29.7          & 37.0                            \\
    Qwen3-Omni-30B-A3B-Thinking~\cite{Qwen3-Omni}       & 26.4          & \underline{37.2}    & 38.5          &  & 46.8          & 35.6          & 35.5          & 35.2          & 37.2                            \\
    \rowcolor[HTML]{E7E6E6} 
    \rowcolor[HTML]{E7E6E6} 
    \method       & \textbf{40.7} & \textbf{38.1} & \textbf{46.6} &  & \textbf{53.8} & \textbf{43.5} & \textbf{43.9} & \textbf{41.4} & \textbf{44.8 \textcolor{red}{(+7.8pp)}}                   \\ \hline
    \end{tabular}
    \caption{Accuracy comparison of \method and other methods on OmniVideoBench~\cite{omnivideobench}. The \textbf{best} is highlighted and the second-best is \underline{underlined}. The performance gains of our method over the base model are indicated in \textcolor{red}{red} parentheses.}\label{table:ovdbench}
\end{table*}

We first assess \method on a suite of audio-video understanding benchmarks. After the \method training phase, the model shows remarkable improvements across multiple benchmarks. Notably, \method outperforms the open-source SOTA model Video-SALMONN 2+-72B (which has a larger parameter scale) by at least \textbf{4.3\%} (82.8 vs. 79.4). Additionally, on specific benchmarks, \method even exceeds the latest closed-source SOTA model Gemini3-Pro, achieving a \textbf{2.1\%} advantage (82.8 vs. 81.1) on Daily-Omni and a \textbf{3.8\%} improvement (74.2 vs. 71.5) on IntentBench.

Interestingly, certain reasoning-oriented variants have been observed to underperform compared to their base counterparts on specific benchmarks (e.g., Qwen3-Omni-30B-A3B-Thinking vs. Qwen3-Omni-30B-A3B shows 48.0 vs. 54.0 on WorldSense). In contrast, \method consistently demonstrates superior performance over the base model across all evaluated benchmarks, underscoring both the effectiveness and robustness of our approach.

Furthermore, we evaluate \method on a more challenging benchmark focused on synergistic audio-visual understanding, with a strong emphasis on modality complementarity and logical consistency. As shown in Tab.~\ref{table:ovdbench}, \method surpasses Qwen3-Omni-30B-A3B by \textbf{21.1\%} (44.8 vs. 37.0). Previous methods performed close to random guessing on this benchmark, but \method breaks through this bottleneck and achieves significant gains, consistently surpassing the base model across all evaluation dimensions. These results highlight the substantial potential of audio-visual joint reasoning through accurately grounded key cues.

\subsection{Visual-only Understanding}
\begin{table*}[t!]
    \centering
    \footnotesize
    \setlength{\tabcolsep}{5.0mm}
    \begin{tabular}{lccc}
    \hline
    \textbf{Method}             & \textbf{Video-MME} & \textbf{MLVU(Dev)} & \textbf{LVBench} \\ \hline
    GPT-4o                      & 71.9               & 64.6               & 30.8             \\
    Gemini-2.0-Flash            & 72.4               & 71.0               & 57.9             \\
    Gemini-2.5-Pro~\cite{google_gemini_25_pro}              & 86.9               & 81.2               & 69.2             \\ \hline
    VideoLLaMA3-7B~\cite{zhang2025videollama3}              & 66.2               & 73.0               & 45.3             \\
    InternVideo2.5-8B~\cite{wang2025internvideo2d5}           & 65.1               & 72.8               & 46.4             \\
    Qwen2.5-VL-7B~\cite{bai2025qwen2.5}               & 65.1               & 70.2               & 45.3             \\
    Qwen2.5-VL-72B~\cite{bai2025qwen2.5}              & 73.3               & 74.6               & 47.3             \\
    video-SALMONN 2+-7B~\cite{video-salmonn-2}         & \underline{73.4}         & 73.6               & 49.7             \\
    Qwen3-Omni-30B-A3B-Instruct~\cite{Qwen3-Omni} & 70.5               & \underline{75.2}         & 50.2             \\
    Qwen3-Omni-30B-A3B-Thinking~\cite{Qwen3-Omni} & 69.7               & 72.9               & 49.0             \\
    \rowcolor[HTML]{E7E6E6} 
    \rowcolor[HTML]{E7E6E6} 
    \method & \textbf{73.6}      & 74.1               & \textbf{51.9}    \\ \hline
    \end{tabular}
    \caption{Performance of different methods on various visual-only benchmarks, including Video-MME~\cite{videomme}, MLVU~\cite{mlvu} and LVBench~\cite{wang2025lvbench}. The \textbf{best} is highlighted and the second-best is \underline{underlined}.}\label{table:visual}
    \vspace{-1em}
\end{table*}

On the other hand, to assess whether the model suffers performance degradation in a single modality after mixed-modality post-training, we evaluate \method on a suite of silent-video benchmarks. As shown in Tab.~\ref{table:visual}, \method exhibits no evident degradation and even demonstrates improvements compared to the base model; specifically, it achieves gains of  \textbf{4.4\%} (73.6 vs. 70.5), \textbf{-1.4\%} (74.1 vs. 75.2), and \textbf{3.4\%} (51.9 vs. 50.2) on Video-MME, MLVU, and LVBench, respectively.

This robustness stems from the model's ability to effectively ground behaviors during inference, allowing it to proficiently capture key cues regardless of whether the input is purely visual or audio-visual. These results confirm our core objective of fostering modality integration to enhance reasoning, rather than resulting in trade-offs between modalities.

\subsection{Different Training Strategies}
\begin{table}[t!]
    \centering
    \footnotesize
    \setlength{\tabcolsep}{1.5mm}
    \begin{tabular}{lccc}
    \hline
    \textbf{Method} & \multicolumn{1}{l}{\textbf{OmniVideoBench}} & \textbf{Daily-Omni} & \textbf{WorldSense} \\ \hline
    QA SFT          & 39.1                                        & 69.9                & 57.4                \\
    CoT SFT         &  42.2                                           & 73.1                    &  59.2                   \\
    Vanilla RL & 41.5                                        & 73.9                & 58.0                \\
    \rowcolor[HTML]{E7E6E6} 
    Full            & \textbf{44.8}                               & \textbf{82.8}       & \textbf{65.8}       \\ \hline
    \end{tabular}
    \caption{Performance on different training strategies in terms of Qwen3-Omni-30B-A3B-Instruct. The \textbf{best} is highlighted.}\label{table:training}
    \vspace{-1em}
\end{table}
Following the dataset $\mathcal{D}$ curated for \method,  we first attempt to use these 88173 examples to directly learn the final response in the QA SFT setting, as reported in Tab.~\ref{table:training}. That is, the model is supervised only on the final answers. In contrast, CoT SFT augments $\mathcal{D}$ with chain-of-thought annotations generated by Gemini-2.5-
Pro~\cite{google_gemini_25_pro} and then fine-tunes the model on these CoT-augmented examples. Vanilla RL instead applies standard GRPO~\cite{rela:deepseekr1} on $\mathcal{D}$ under a \texttt{<think>...</think>}\texttt{<answer>...</answer>} protocol, using a mixture of format and soft-response scores as the reward. As shown in Tab.~\ref{table:training}, all these approaches yield noticeable improvements over the base model on audio–visual understanding benchmarks, \emph{confirming the effectiveness of $\mathcal{D}$ after our data preparation pipeline.}

However, the performance gains of these baselines are consistently smaller than those achieved by \method. On Daily-Omni, our method surpasses the second-best Vanilla RL by \textbf{12.0\%} (82.8 vs. 73.9), and on WorldSense it outperforms the second-best CoT SFT by \textbf{11.1\%} (65.8 vs. 59.2). \emph{These ablation results further validate the effectiveness and superiority of our training paradigm.}



\subsection{Case study}\label{sec:case}
\begin{figure*}[!htbp]
    \centering
    \includegraphics[width=0.88\textwidth]{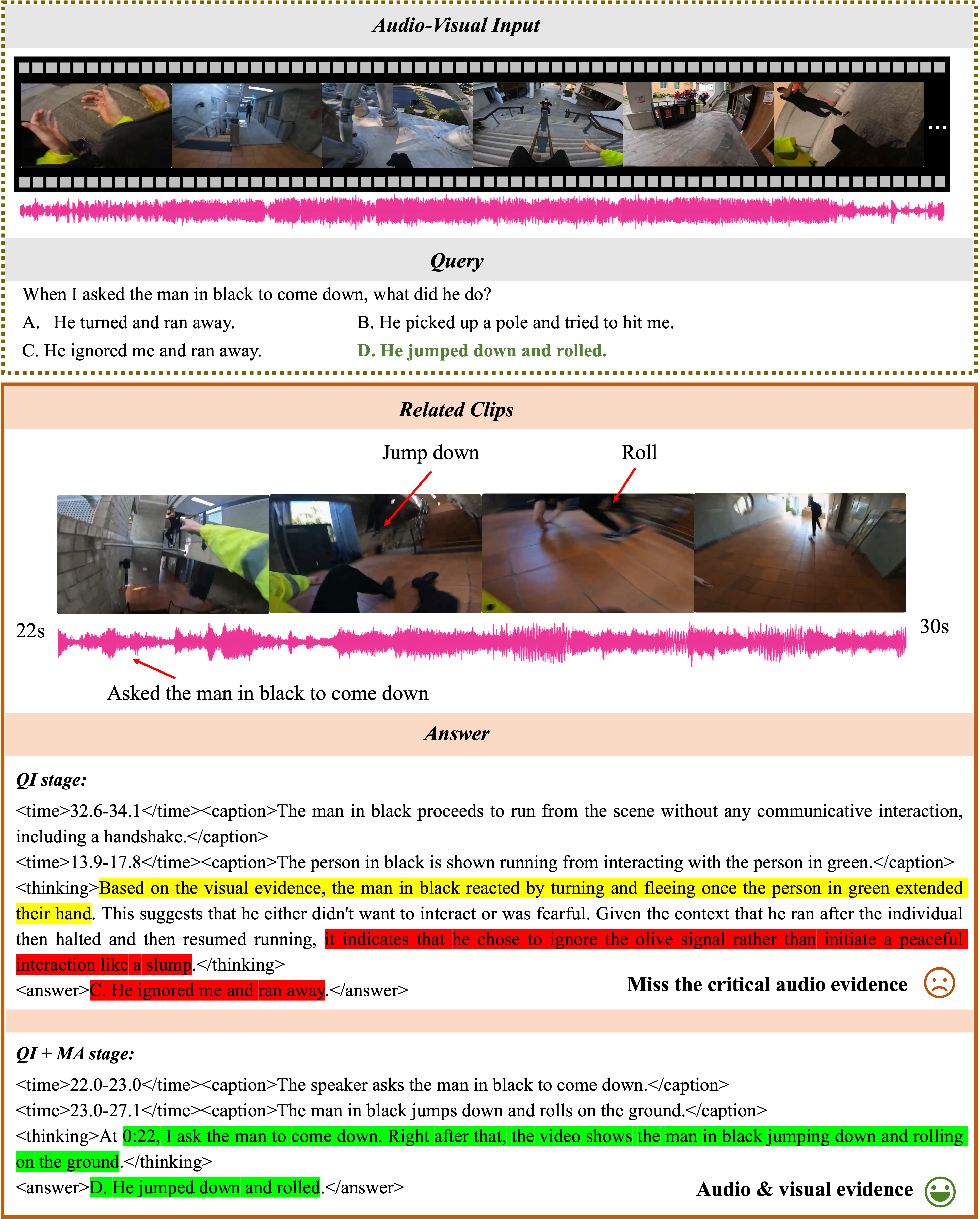}
    \caption{Visualization of the results obtained from the training of QI, and QI+MA. \textcolor{red}{Red} highlights the incorrect text, while \textcolor{green}{green} highlights the correct text. \textcolor{yellow}{Yellow} highlights the model overemphasizes one modality while neglecting cues from the other modality.}\label{fig:show1}
\end{figure*}

We further present several qualitative cases for QI-only training, and our QI+MA (\method) training. As shown in Fig.~\ref{fig:show1}, QI training yields strong reasoning behavior; however, in some cases the model overlooks critical audio cues, resulting in inaccurate inferences. In contrast, our QI+MA, first establishing the desired reasoning behavior and then booming deeper audio-visual reasoning, enables the model to better exploit both audio and visual evidence.

\begin{figure*}[!htbp]
    \centering
    \includegraphics[width=0.62\textwidth]{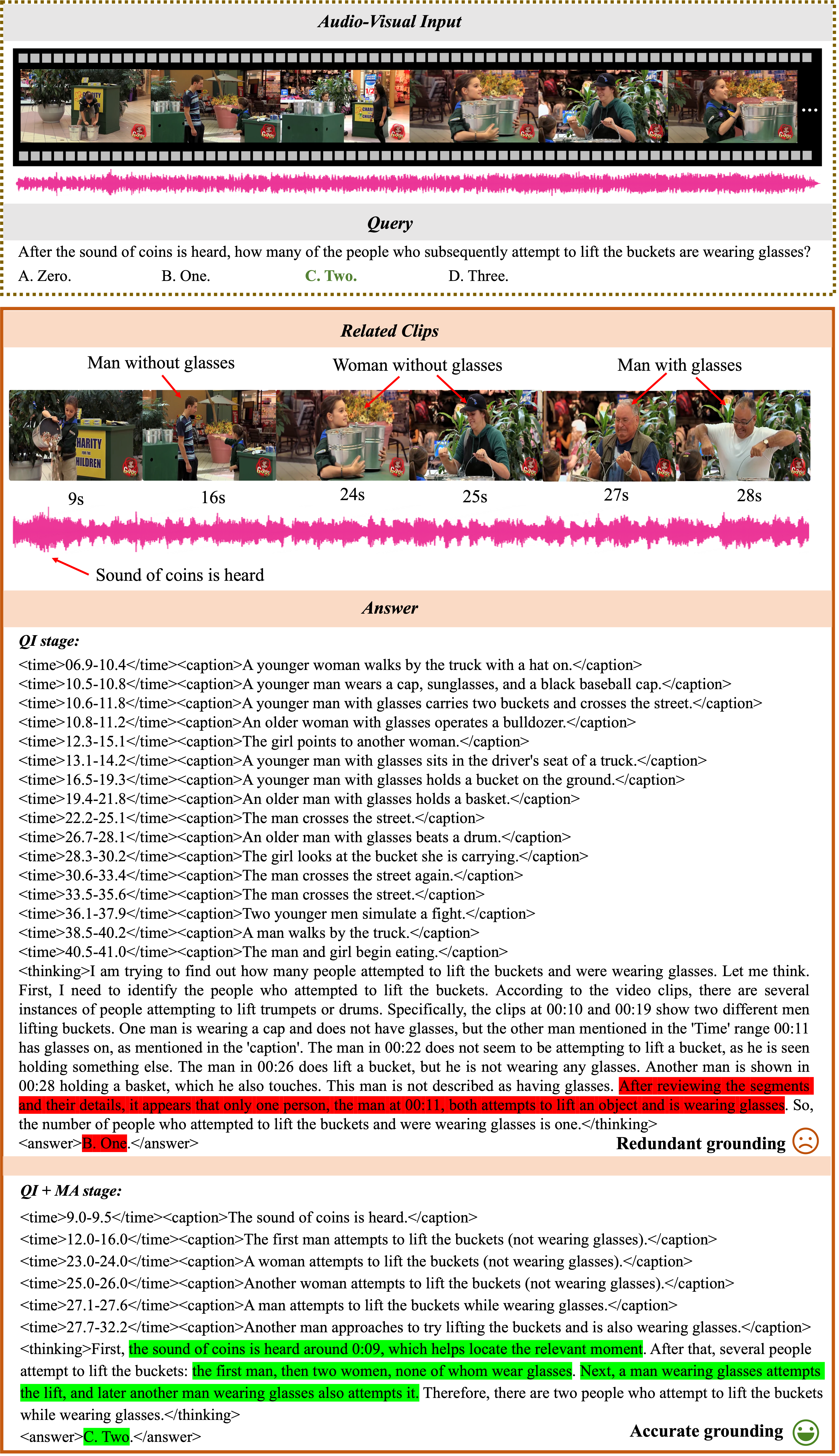}
    \caption{Visualization of the results obtained from the training of QI, and QI+MA. \textcolor{red}{Red} highlights the incorrect text, while \textcolor{green}{green} highlights the correct text.}\label{fig:show2}
\end{figure*}

Moreover, as illustrated in Fig.~\ref{fig:show2}, QI training tends to introduce redundant grounding, as its primary objective is to shape reasoning behavior. Our QI+MA further use MA to maximize the utilization of audio-visual cues.

\subsection{Ablation Study}
\begin{table}[t!]
    \centering
    \setlength{\tabcolsep}{0.5mm}
    \footnotesize
    \begin{tabular}{lccc}
    \hline
    \textbf{Method}       & \multicolumn{1}{l}{\textbf{OmniVideoBench}} & \textbf{Daily-Omni} & \textbf{WorldSense} \\ \hline
    w/o $r_{\mathrm{intent}}$         & 38.4                                        & 75.9                & 55.1                \\
    w/o $r_{\mathrm{attn}}$     & 43.7                                        & 82.1                & 65.5                \\
    w/o QI (10K)      & 41.6                                        & 76.1                & 58.6                \\
    w/o QI (80K)      & 41.0                                        & 76.9                & 59.2                \\
    w/o MA     & 43.6                                        & 82.0                & 65.3                \\
    w/o audio input & 37.6                                        & 68.7                & 50.3                \\
    w. timestamps              & 43.4                                        & 81.7                & \textbf{66.1}       \\
    Base model       & 37.0                                        & 63.6                & 54.0                \\
    \rowcolor[HTML]{E7E6E6} 
    Full                  & \textbf{44.8}                               & \textbf{82.8}       & 65.8                \\ \hline
    \end{tabular}
    \caption{Performance on different ablated settings in terms of Qwen3-Omni-30B-A3B-Instruct. The \textbf{best} is highlighted.}\label{table:ablation}
    \vspace{-2.5em}
\end{table}
\noindent\textbf{Component Removal.}
We first perform ablations on various designs (w/o $r_{\mathrm{attn}}$, $r_{\mathrm{intent}}$, or QI stage) in Tab.~\ref{table:ablation}. The results suggest that the observed performance gains are mainly attributable to two factors: \emph{(1) $r_{\mathrm{intent}}$, which encourages accurate grounding of the primary cues, and (2) modality-attentive training, which further strengthens the model's ability to perform comprehensive audio-visual reasoning.}

It can be observed that \emph{performing MA stage training alone can bring substantial improvements} (as shown in ``w/o QI'' setting in the table). For instance, on OmniVideoBench, this strategy yields a \textbf{12.4\%} gain over the base model (41.6 vs.\ 37.0). Furthermore,  
QI stage training (``w/o MA'' setting in the table) also significantly improved the model's capability, yielding a \textbf{17.8\%} gain over the base model (43.6 vs.\ 37.0). Removing $r_{\mathrm{intent}}$ or $r_{\mathrm{attn}}$ both results in certain performance drop.


\noindent\textbf{Input Configuration.}
We further ablate the impact of input configuration in Tab.~\ref{table:ablation}. Specifically, we use \method trained with both audio and video, but perform inference under a ``w/o audio input'' setting. We observe a performance drop on WorldSense (50.3 vs. 54.0), but a slight improvement on Daily-Omni (68.7 vs. 63.6). This mixed behavior can be attributed to two factors: (1) when the evaluation benchmark inherently relies on audio, the \emph{mismatch} between training (w. audio) and inference (w/o audio) naturally leads to degraded performance; (2) owing to the enhanced reasoning capability and \emph{robust grounding of key visual cues}, the model can actually \emph{perform better on tasks where audio is non-essential or visual information alone is sufficient.}

Moreover, many recent methods introduce explicit temporal cues by overlaying timestamps~\cite{ge2025arc}. While this can strengthen temporal perception, it simultaneously \emph{occludes part of the original visual content}. In contrast, our $r_{\mathrm{intent}}$ reward inherently promotes temporal correction during training (e.g., \emph{inaccurate temporal grounding directly degrades caption quality}), endowing \method with an implicit sense of time. As a result, our method is insensitive to such numeric overlays (``w. timestamps''), exhibiting only marginal performance differences.

\section{Conclusion}
\label{sec:conclusion}
In this paper, we propose \method, a query-intensive deep fusion framework for audio-visual reasoning. Our training pipeline consists of two stages. First, without relying on any process-level annotations, we encourage the model to “think with omnimodal cues” by learning in a self-supervised manner grounded in intermediate time-caption pairs. Second, we explicitly enhance cross-modal fusion by contrasting the model’s learning under full audio-visual input versus single-modality input, thereby improving its ability to build coherent multimodal representations. Extensive experiments show that \method consistently outperforms prior methods on multiple benchmarks, laying a solid foundation for future work in audio-visual reasoning.


\nocite{langley00}

\bibliography{main}
\bibliographystyle{icml2026}
\clearpage
\newpage
\appendix
In this Appendix, we provide more technical details, including 1) Detailed descriptions of training dataset and benchmarks in Sec.~\ref{sec:data}; 2) the specific prompts used in our experiments in Sec.~\ref{sec:instruction}; 3) the hyperparameter configurations of training settings in Sec.~\ref{sec:implementation}; and 4) the limaitation and future work in Sec.~\ref{sec:limit}

\section{Dataset}\label{sec:data}
\label{sup:dataset}
\subsection{Training Dataset}
\begin{figure*}[!htbp]
    \centering
    \includegraphics[width=0.7\textwidth]{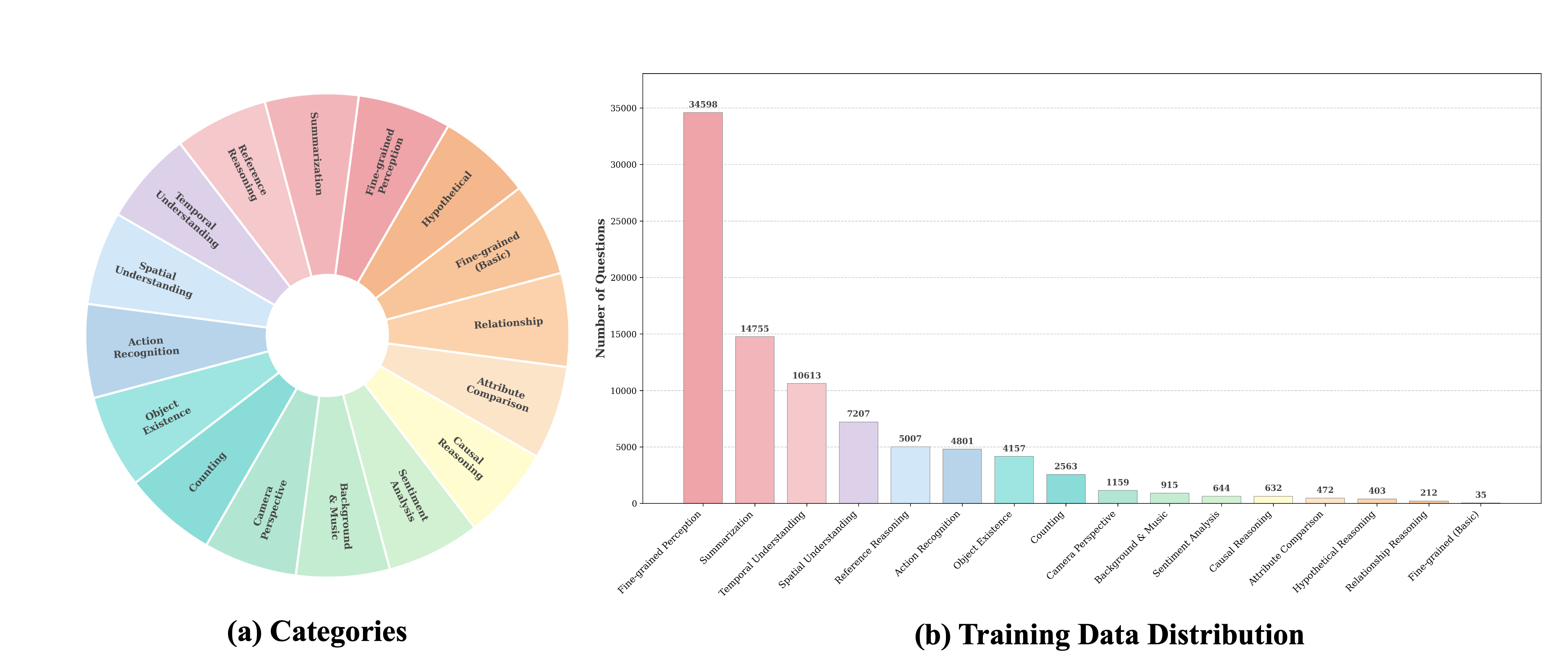}
    \caption{(a) Our training data covers 16 categories. (b) Number of questions in terms of each category.}\label{fig:supdata}
\end{figure*}
As illustrated in Fig.~\ref{fig:supdatappl}, we primarily perform a three-stage data filtering and selection pipeline to obtain high-quality audio–video data. To more clearly present the distribution of the processed data, we report descriptive statistics of the resulting training dataset as shown in Fig.~\ref{fig:supdata}. That is, our dataset comprises 16 categories with varying numbers of samples, ranging from 35 to 34598. The questions with audio–video are of high quality and exhibit substantial diversity in content.
\subsection{Benchmarks}
\subsubsection{Audio-visual Benchmarks}
\noindent\textbf{OmniVideoBench~\cite{omnivideobench}:} 
a large-scale, carefully curated benchmark for evaluating synergistic audio–visual reasoning, with particular emphasis on \emph{modality complementarity and logical coherence}. It contains 1000 high-quality question–answer pairs, derived from 628 diverse videos spanning from a few seconds to 30 minutes.

\noindent\textbf{Daily-Omni~\cite{dailyomni}:} 
an audio–visual question answering dataset containing 684 \emph{daily-life videos} from diverse sources, rich in both auditory and visual cues, and providing 1197 multiple-choice QA pairs spanning 6 major tasks.

\noindent\textbf{WorldSense~\cite{worldsense}:} 
a benchmark emphasizing \emph{omnimodal collaboration}, with strongly coupled audio–video tasks that require synergistic multimodal perception. It contains 1662 synchronized audio–visual videos across 8 domains and 67 subcategories, and 3172 multiple-choice QA pairs covering 26 tasks for comprehensive evaluation

\noindent\textbf{IntentBench~\cite{yang2025humanomniv2}:} a benchmark designed to evaluate models’ understanding of complex \emph{human intentions and emotions}, comprising 633 videos and 2689 questions grounded in both auditory and visual cues.

\noindent\textbf{VideoHolmes~\cite{videoholmes}:} a Sherlock Holmes–inspired benchmark for evaluating \emph{complex video reasoning} in MLLMs, featuring 1837 questions from 270 annotated suspense short films across seven tasks, each requiring models to connect dispersed visual clues and underlying causal events.

\subsubsection{Visual-only Benchmarks}
\noindent\textbf{Video-MME~\cite{videomme}:} the first \emph{full-spectrum, multimodal evaluation benchmark} for MLLMs in video analysis, covering 6 major visual domains and 30 subdomains, with 900 videos ranging from 11 seconds to 1 hour (totaling 254 hours) and 2700 QA pairs.

\noindent\textbf{MLVU~\cite{mlvu}:} the benchmark focuses on \emph{long videos and diversity in both video types and evaluation tasks}, with durations ranging from 3 minutes to 2 hours and a total of 9 different evaluation tasks. In this paper, \emph{we use its dev subset for evaluation.}

\noindent\textbf{LVBench~\cite{wang2025lvbench}:} a benchmark specifically designed for \emph{ultra-long video understanding spanning several hours}, aimed at testing MLLMs’ long-term memory and extended comprehension abilities. It contains 103 videos and 1549 question–answer pairs in total.

\section{Instruction Details}\label{sec:instruction}
\subsection{\method}
\begin{figure*}[]
    \centering
    \includegraphics[width=0.7\textwidth]{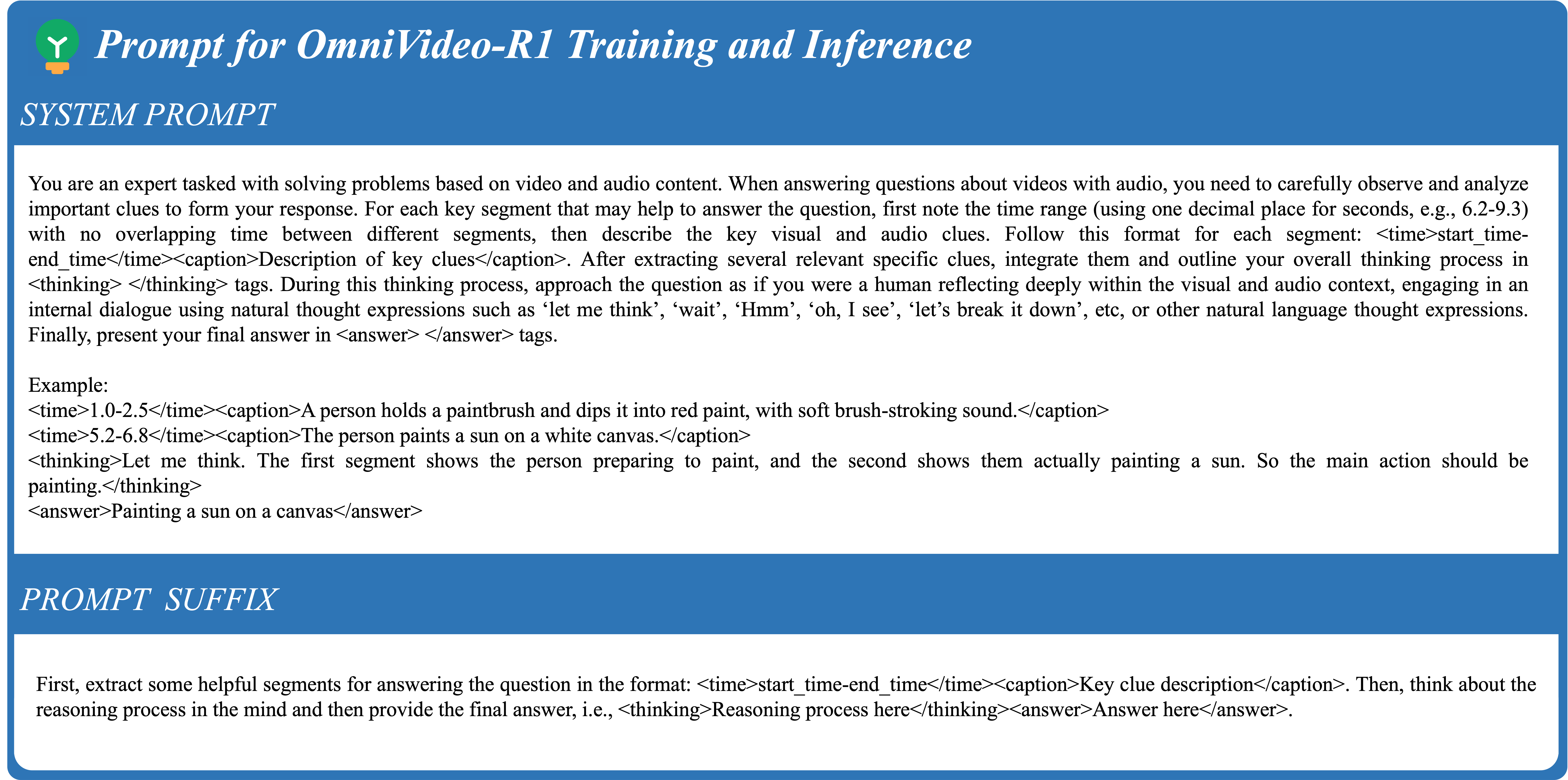}
    \caption{System prompt and user prompt suffix for \method reasoning.}\label{fig:methodprompt}
\end{figure*}
As illustrated in Fig.~\ref{fig:methodprompt}, we employ a specialized system prompt as well as the fixed suffix to the user prompt for \method. In this way, the model can undergo training for specific reasoning behaviors starting from zero-RL.

\subsection{Data Preparation}
\begin{figure*}[]
    \centering
    \includegraphics[width=0.7\textwidth]{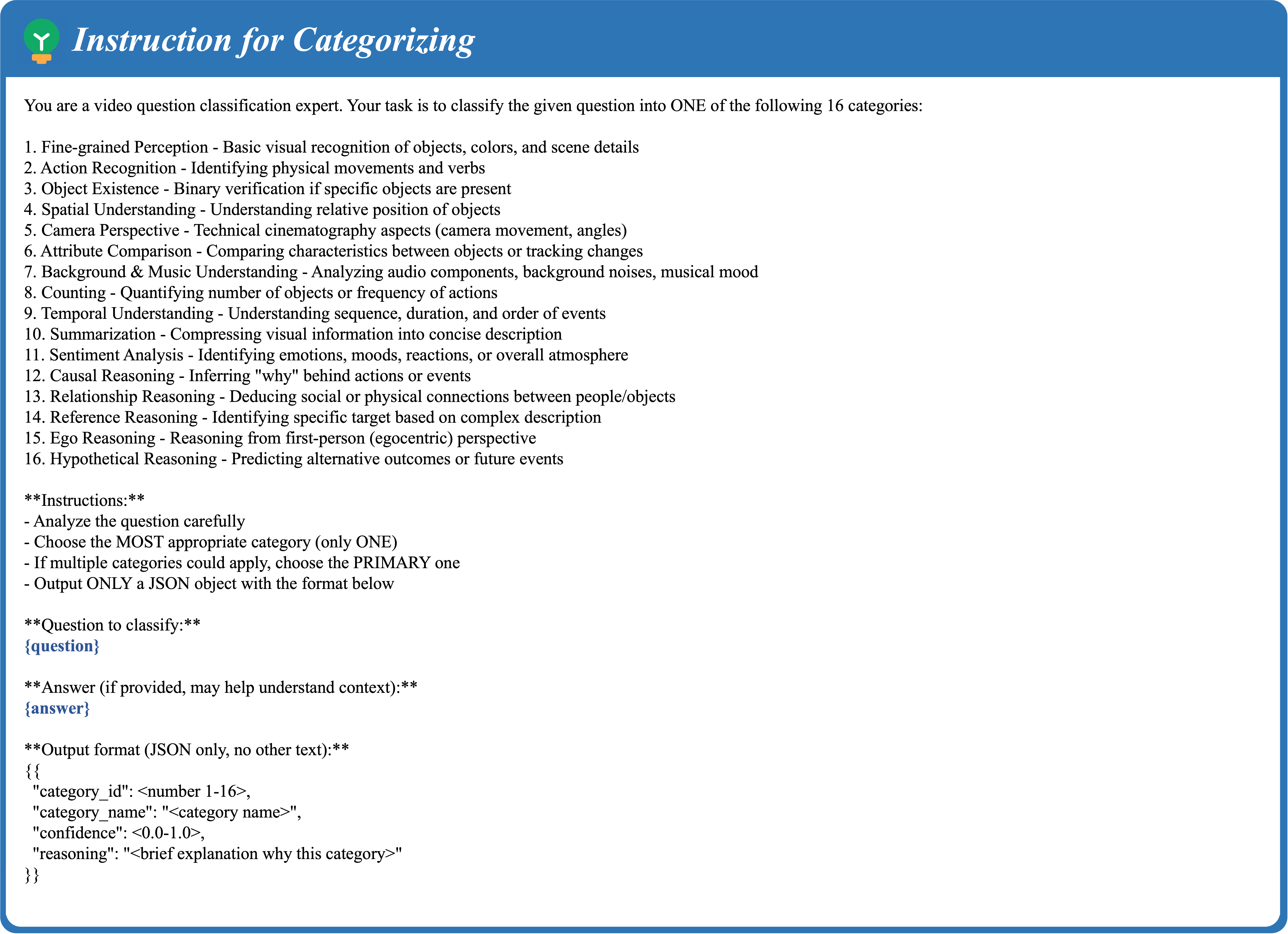}
    \caption{Instruction for data categorizing in data preparation.}\label{fig:categoryprompt}
\end{figure*}
\begin{figure*}[!htbp]
    \centering
    \includegraphics[width=0.9\textwidth]{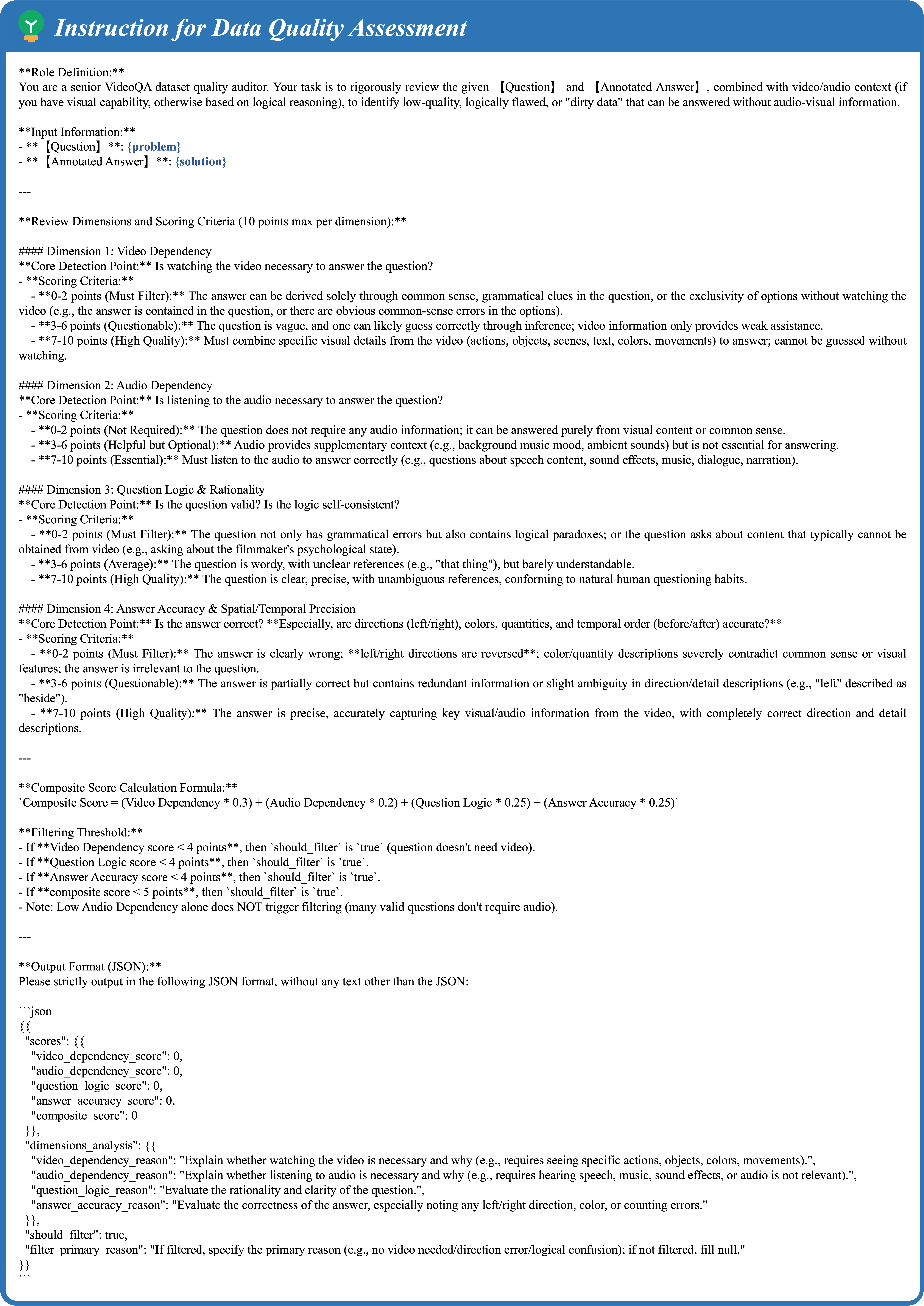}
    \caption{Instruction for data quality assessment in data preparation.}\label{fig:scoreprompt}
\end{figure*}
\begin{figure*}[]
    \centering
    \includegraphics[width=0.9\textwidth]{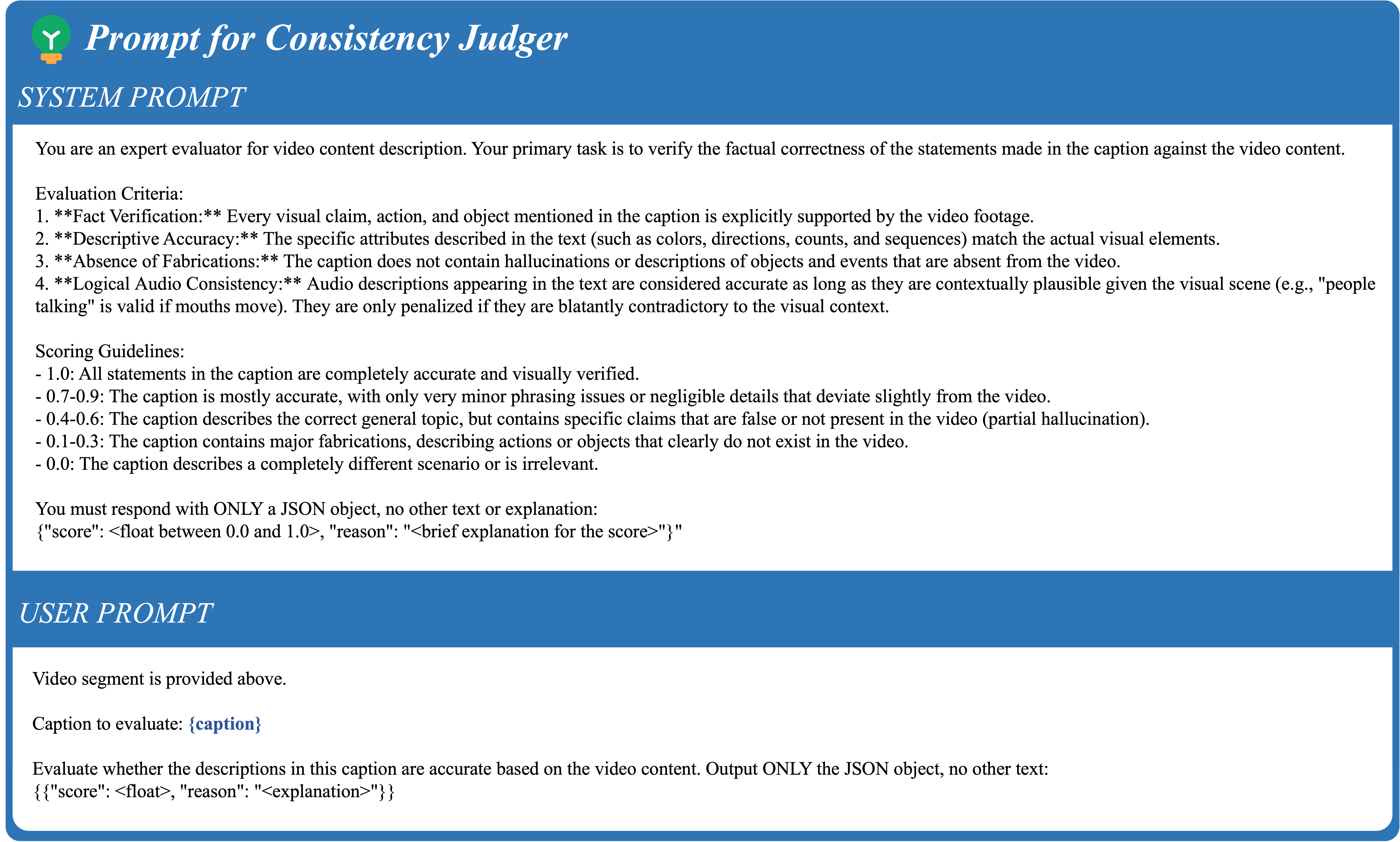}
    \caption{System prompt and user prompt for consistency judger.}\label{fig:consistencyprompt}
\end{figure*}
\begin{figure*}[]
    \centering
    \includegraphics[width=0.9\textwidth]{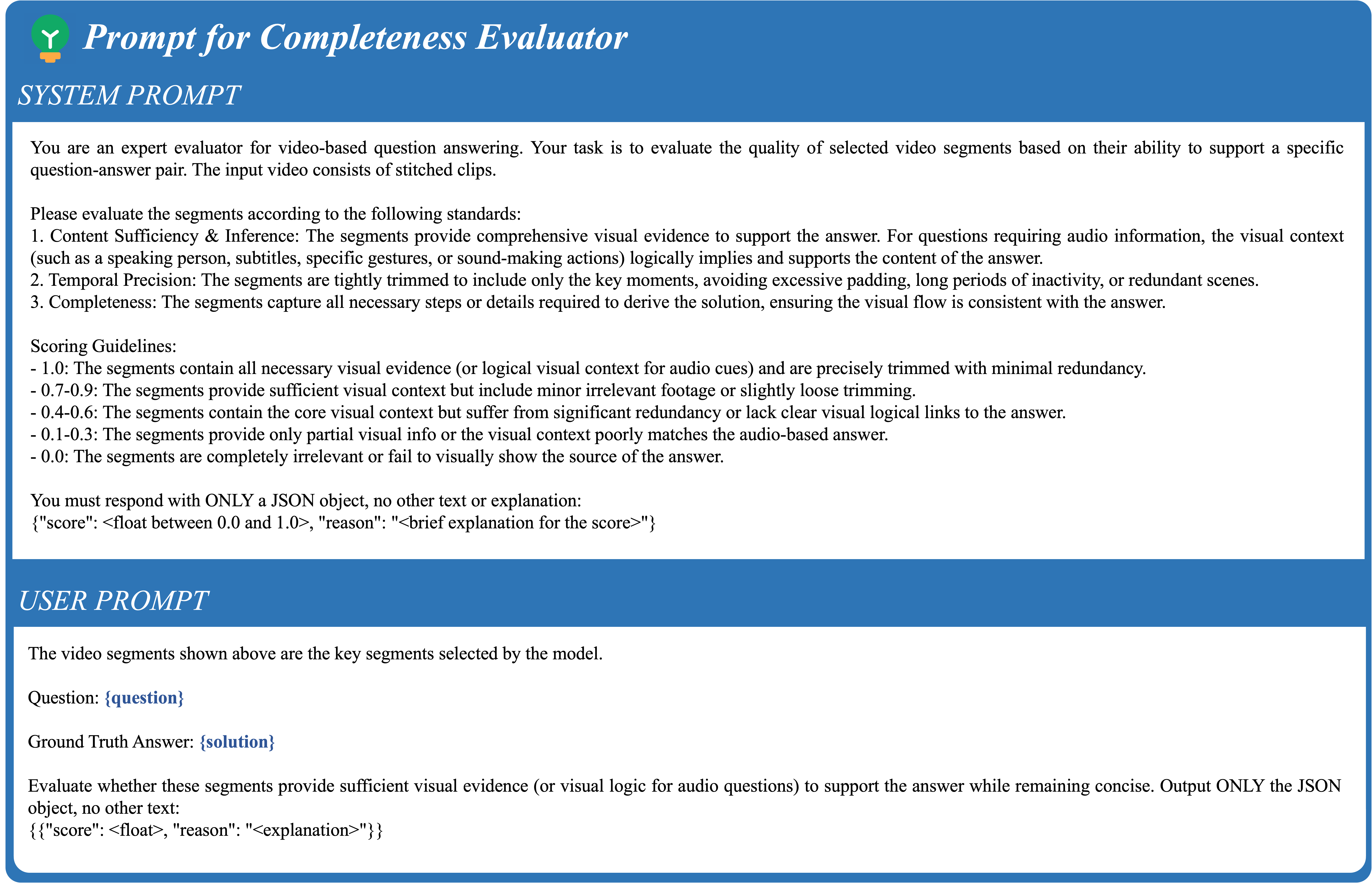}
    \caption{System prompt and user prompt for completeness evaluator.}\label{fig:completenessprompt}
\end{figure*}
In terms of data preparation, in the last stage, we perform categorization based on Qwen-3-32B~\cite{yang2025qwen3} as the instruction shown in Fig.~\ref{fig:categoryprompt}, dividing the data into 16 categories. The final results of this categorical analysis are shown in Fig.~\ref{fig:supdata}.

\subsection{Consistency Judger}
As shown in Fig.~\ref{fig:pipeline}, the consistency judger is mainly used for rewarding the consistency of time-caption pairs. It is primarily based on Qwen3-VL-235B-A22B-Instruct~\cite{Qwen3-VL} for scoring, and the corresponding prompt is illustrated in Fig.~\ref{fig:consistencyprompt}.

\subsection{Completeness Evaluator}
Meanwhile, as shown in Fig.~\ref{fig:pipeline}, the completeness evaluator is mainly used to assess the completeness of multiple audio–video segments produced by query-intensive grounding. It is also based on Qwen3-VL-235B-A22B-Instruct~\cite{Qwen3-VL} for scoring, and the corresponding prompt is illustrated in Fig.~\ref{fig:completenessprompt}.

\section{Implementation details}\label{sec:implementation}
We set more of the key hyperparameters as follows: FPS\_MAX\_FRAMES 64 to cap the number of frames per sample, lr\_warmup\_fraction 0.05 to gradually ramp up the learning rate at the start of training, $\epsilon$ $3\times10^{-4}$ and $\epsilon_{\text{high}}$ $4\times10^{-4}$ as clipping thresholds, KL regularization coefficient $\beta$ 0.03 to penalize large deviations from the reference policy, and moe\_aux\_loss\_coeff $10^{-3}$ to weight the auxiliary load-balancing loss for the mixture-of-experts.

\section{Limitation \& Future Work.}\label{sec:limit}
(1) Current methods still rely on outcome-based ground-truth for training. Exploring how to effectively strengthen the model in the absence of ground-truth could be an important direction for future research. (2) The multimodal training paradigm is not restricted to audio–visual inputs. With query intention and modality attention, it can extend to more modalities (e.g., 3D).
\end{document}